# Image Captioning with Object Detection and Localization

Zhongliang Yang, Yu-Jin Zhang, Sadaqat ur Rehman, Yongfeng Huang,

Department of Electronic Engineering, Tsinghua University, Beijing 100084, China
Email: yangzl15@mails.tsinghua.edu.cn

**Abstract.** Automatically generating a natural language description of an image is a task close to the heart of image understanding. In this paper, we present a multi-model neural network method closely related to the human visual system that automatically learns to describe the content of images. Our model consists of two sub-models: an object detection and localization model, which extract the information of objects and their spatial relationship in images respectively; Besides, a deep recurrent neural network (RNN) based on long short-term memory (LSTM) units with attention mechanism for sentences generation. Each word of the description will be automatically aligned to different objects of the input image when it is generated. This is similar to the attention mechanism of the human visual system. Experimental results on the COCO dataset showcase the merit of the proposed method, which outperform previous benchmark models.

## 1 Introduction

In the past few years, deep neural network has made significant progress in image processing area, like image classification [1, 2, 3], object detection [4, 5, 6]. However, tasks like image classification and object detection are far from the end of image understanding. One ultimate goal of image processing is deep image understanding to machines i.e. understanding the whole image scenario not individual objects. Image captioning follows the same path: by extracting the complete detail of individual object and their associated relationship from image. Finally, the system can automatically generate a neural sentence to describe the image. This problem is extremely important, as well as difficult because it connects two major artificial intelligence fields: computer vision and natural language processing.

Previous studies are mostly inspired by the works of neural machine translation, where the task is to translate a source sentence written in one language (like French) into a target sentence written in a different language (like English), keeping logic and syntax precise. Associating image captioning and machine translation together makes sense because they can be placed in the same framework, called Encoder-Decoder framework. In the Encoding step, we encode the source information, which is an image in image captioning task and source sentence in translation task, into a target vec-tor. Followed by the decoding step, in which sentences are generated by decoding the target vector. The core part of this framework is how to encode the source information (image or sentence) and how to decode the target vectors.

Previous works are align in the decoding step, they usually use recurrent neural networks (RNN) based on long short-term memory (LSTM) [7] units as a decoder. As for the encoding step is concern, the work is divided in to two major classes: CNN-RNN models [8, 9, 10] and attention based models [11, 12]. CNN-RNN models represent an image as a single feature vector from the top layer of a pre-trained convolutional network whereas, attention based models use the vector made up by the representations of image's subregions as the source vector. The biggest drawback of CNN-RNN models is that they hardly align different visual parts of the input image to words in captions. Attention based model like [11] allowed their model to attend any visual parts of the input image but most of the subregions it attend are meaningless.

Our model also follows the Encoder-Decoder framework, however, it is totally different from the previous models as shown in figure 1. Our motivation in describing an image is to find out its contents, rather than focusing on some meaningless regions associated with it. As the locations of different objects in an image is also a positive information for describing an image, since they reflect the spatial position relationships of objects in an image. Due to the aforementioned reason, we combine object detection with image captioning to focus on the real meaningful information domain in the images to generate the resulting sentences much better and efficient. The encoding part of our model consists of two steps. Initially, we use an object detection model to detect objects in the image followed by a deep Convolutional Neural Network to extract their spatial relationships. All these information's will be represented as a set of feature vectors, which is then fed into the decoding part where the description sentence will be generated.

In order to measure the performance, we evaluate our model on COCO dataset using seven standard metrics: BELU-1,2,3,4, METEOR, CIDEr and ROUGH-L. Experimental results shows that our proposed model perform better than the baseline soft attention model [11] and is similar to the benchmark ATT model [12] in performance evaluation.

## 2  Our Model

This section describe the detail of the proposed model, which consists of two main parts: encoding and decoding. The input to our model is a single image $I$, while the output is a descriptive sentence $S$ consists of $K$ encoded words: $S = \{w_1, w_2, ..., w_K\}$.

In the encoding part, firstly, we present a model that recognizes objects in the input image followed by a deep CNN to extract their locations, which reflect the spatial relationship associated. All the information will be represented as a set of feature vectors referred as annotation vectors. The encoding part produces $L$ annotation vectors, each of which is a D-dimensional representation corresponding to an object and also its spatial location in the input image: $A = \{A_1, A_2, ..., A_L\}, A_i \in R^D$.

In the decoding part, all these annotation vectors are fed into a deep Recurrent Neural Network model to generate a description sentence.

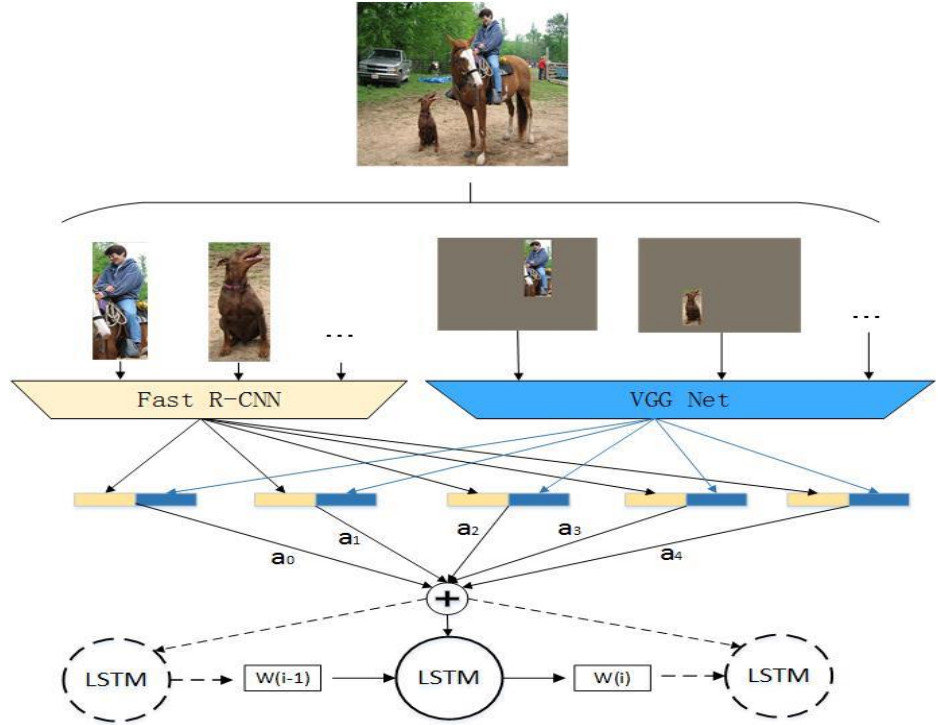

**Fig. 1.** An overview of the proposed framework. The encoding part first extract the information of objects(left) and their spatial relationships(right) in the image, then the decoding part generate words based on these features.

### 2.1 Encoding part

Our core insight is that when human beings try to describe an image using sentence (combination of words), it's natural to first find out objects and their relationships in the desired image. To imitate human beings, our encoding part has two steps, first we use an object detection model to detect objects in the image followed by a deep Convolutional Neural Network to get their spatial locations.

**Object Detection.** In the past few years, significant progress have been done in object detection. These advances are driven by the success of region proposal methods (e.g. [13]) and region-based convolutional neural networks(R-CNN) [5]. In our model, we choose Faster R-CNN [6] as object detection model due to its efficiency and effectiveness in object detection task. Faster R-CNN is composed of two modules. The first module is a deep fully convolutional network that propose regions, the second module is the Fast R-CNN detector [5] which uses the proposed regions. To generate

44

region proposals, the authors of [6] slides a small network over the convolutional feature map output by the last shared convolutional layer. For each sliding window of the input convolutional feature map, this small network maps it to a lower-dimensional feature. Then this feature fed into two sibling fully-connected layers: a box-regression layer (reg) and a box-classification layer(cls). After training, the Faster R-CNN takes an image (of any size) as an input and produce a set of rectangular object proposals, each with an objectness score. Then we sort these boxes according to their scores in descending order and choose the *top-n* boxes as the regions of objects in the input image. Each rectangular object region is mapped to a feature vector by fully connected layers (in our implementation it's 'fc7' layer) of the Fast R-CNN model. More explicitly, for every input image we detect $n$ objects in an image and each object is represented as a d-dimension vector: { $obj_1, obj_2, ..., obj_n$}, $obj_i \in d^d$.

**Object Localization.** This part is designed to extract the information of objects spatial locations which in turn reflect their spatial relationships. Junqi *etal.* [14] also considered the locations of different localized regions. However, they just added the boxes central's with x location, y location, width, height and area ratio with respect to the entire image's geometry to the end of the vector of each localized regions. In this paper, the implementation of extracting information of each object location is completely different from [14]. In object detection part, we know that for each input image, the output is $n$ rectangular object regions, each with an objectness score. For each object in this image, we keep the region of its bounding box unchanged and set remaining regions to mean value of the training set. So we get a new image, which has the same size as the original image but just consists the bounding box region of one object as shown in Figure 1. As we detected $n$ objects for each image therefore, we get $n$ new images for individual image. Each new image will then be fed into the VGG net [2] and the feature vector of its 'fc7' layer will be extracted, which yields to the vectorized representation of object location. Furthermore, we get another $n$ vectors of t-dimension in which each vector represents the information of spatial location of each object: { $loc_1, loc_2, ..., loc_n$}, $loc_i \in d^t$.

Each annotation vector $A_i$ consists of two parts: First, vector $obj_i$ represents the feature of object which particularly describes the contents of image. Second, vector $loc_i$ represents the feature of object location which tell us about the location of individual object.

$$\mathbf{A}_i = [\mathbf{obj}_i\,;\,\mathbf{loc}_i], \mathbf{A}_i \in d^D, D = d + t. \qquad (1)$$

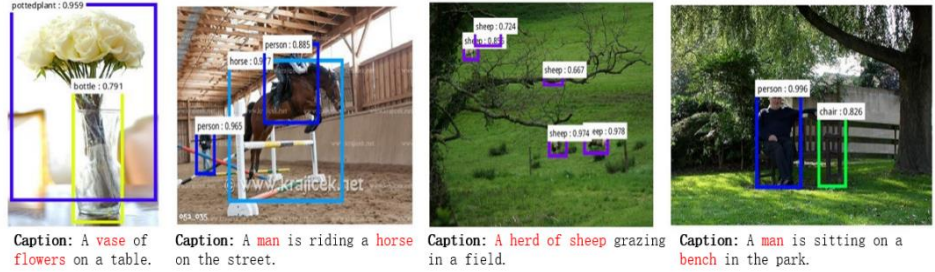

**Fig. 2.** Some results of object detection part. The captions are generated by the proposed model.

### 2.2 Decoding part

In this paper, we describe a decoding part based on an LSTM network with attention mechanism. Attention mechanism was first used in neural machine translation area by [15]. Following the same mechanism, the authors of [16,17,11] introduced it into image processing domain whereas, [11] was the first to apply it in image captioning task. The key idea of attention mechanism is that when a sentence is used to describe an image, not every word in the sentence is "translated" from the whole image but actually it just has relation to a few subregions of an image. It can be viewed as a form of alignment from words of the sentence to subregions of the image. The feature vectors of these subregions are referred to as annotation vectors. Here in our implementation, subregions are referred to as the bounding box of objects and annotation vectors are referred to as $\{A_i\}$, which is already discussed in the encoding part.

In the decoding part we follow [11] to use a long short-term memory(LSTM) network [7] as a decoder. LSTM network products one word at every step $j$ conditioned on a context vector $z_j$, the previous hidden state $h_{j-1}$ and the previously generated words $w_{j-1}$ using the following formulations:

$$In_j = \sigma(W_i E w_{j-1} + U_i h_{j-1} + Z_i z_j + b_i) \qquad (2)$$

$$f_j = \sigma(W_f E w_{j-1} + U_f h_{j-1} + Z_f z_j + b_f) \qquad (3)$$

$$c_j = f_j c_{j-1} + tanh(W_c E w_{j-1} + U_c h_{j-1} + Z_c z_j + b_c) \qquad (4)$$

$$o_j = \sigma(W_o E w_{j-1} + U_o h_{j-1} + Z_o z_j + b_o) \qquad (5)$$

$$h_j = o_j tanh(c_j) \qquad (6)$$

Here $In_j, f_j, c_j, o_j, h_j$ represent the state of input gate, forget gate, cell, output gate and hidden layer respectively. $W, U, Z$ and $b$ are learned weight matrices and biases. E is an embedding matrix and σ is the logistic sigmoid activation. The context vector $z_j$ is generated from the annotation vectors $A_j, i = 1, ..., n$, corresponding to the feature vectors of different objects.

There are two different versions in [11] to compute the context vector $z_j$ and we use the "soft" version, that is

$$\mathbf{z}_j = \sum_{i=1}^{n} \alpha_{ji} \mathbf{A}_i \qquad (7)$$

Where $\alpha_{ji}$ is a scalar weighting of annotation vector $\mathbf{A}_i$ at time step $j$, defined as follows:

$$\mathbf{e}_{ji} = f_{att}(\mathbf{A}_i, h_{j-1}) \qquad (8)$$

$$\alpha_{ji} = \frac{\exp(\mathbf{e}_{ji})}{\sum_{k=1}^{n} \exp(\mathbf{e}_{jk})} \qquad (9)$$

$$\sum_{i=1}^{n} \alpha_{ji} = 1 \qquad (20)$$

Where $f_{att}$ is a multilayer perceptron conditioned on the previous hidden state $h_{j-1}$. The positive weight $\alpha_{ji}$ can be viewed as the probability that the word generated at time step j "translated" from object $i$.

We predict the next word $W_j$ with a softmax layer, the input of it are the context vector, the previously generated word and the decoder state $h_j$:

$$p(w_j) \propto \exp(L_o(Ew_{j-1} + L_h h_j + L_z z_j)) \qquad (31)$$

Where $L_o, E, L_h, L_z$ are learned parameters.

## 2.3 Training

This section describe the training of proposed model. The training data for each image consists of input image features $\{A_i\}$ and output caption words sequence $\{w_k\}$. Parameters of the proposed encoding part is fixed, so we only need to learn the parameters of the proposed decoding part, which are all the attention model parameters $\Theta_{Att} = \{W, U, Z, b\}$ jointly with RNN parameters $\Theta_{RNN}$.

We train our model using maximum likelihood with a regularization term on the attention weights by minimizing a loss function over training set. The loss function is a negative log probability of the ground truth words $w = \{w_1, w_2, \ldots, w_K\}$:

$$LOSS = -\sum_t \log\left(p(w_j)\right) + \lambda \sum_i (1 - \sum_t \alpha_{i,j})^2 \qquad (42)$$

Where $w_j$ is the ground truth word and $\lambda > 0$ is a balancing factor between the cross entropy loss and a penalty on the attention weights. We use stochastic gradient descent with momentum 0.9 to train the parameters of our network.

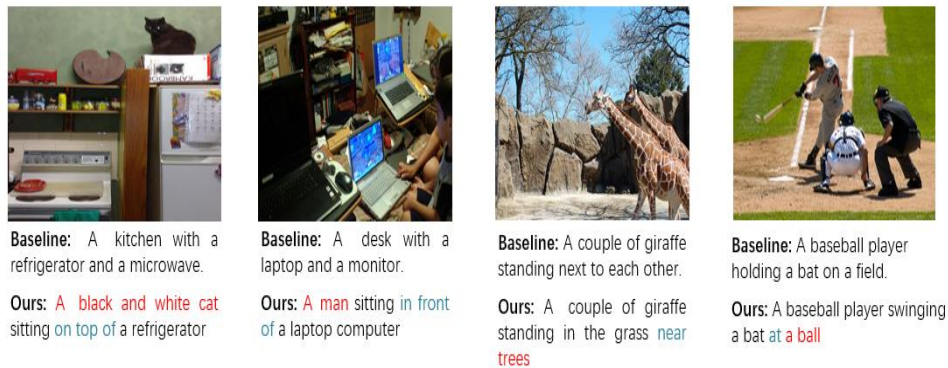

**Fig. 3.** Illustration of how the proposed model use the information of objects and their spatial relationships in the image to generate sentences. The baseline results are generated by Neural-Talk2 (version2.0 of Deep VS [10]). "Ours" are generated by the proposed model. Words in red align to the objects which are not recognized by the baseline model and the blue words show objects spatial relationships.

## 3 Experiments

In this section, we first describe the dataset used in our experiments as well as the experimental methodology followed by the detail results discussion. We report all the results using Microsoft COCO caption evaluation tool [18], including BLEU-1,2,3,4, METEOR, CIDEr and ROUGH-L.

### 3.1 Dataset

We use COCO dataset [22] for experimental purpose to show the efficiency of proposed method, which includes 123287 color images. For each image, there are at least five captions given by different AMT workers. To make the results comparable to other methodologies, we use the commonly adopted [18] splits of the dataset, which assigns 5000 for validation, 5000 for testing and keep the rest for training. For each sentence, we limit the length size to 50 words and truncate the rest. Each sentence ends with a character "<*end*>". Words that appear less than five times are marked as a character "<*unk*>". When dictionary is build, resulting in the final vocabulary with 10020 unique words in COCO dataset.

### 3.2 Experimental Setting

In the encoding part of the proposed model, we use faster R-CNN [5] for object detection and VGG net [2] pre-trained on ImageNet dataset for feature extraction. Faster R-CNN has been trained for the same set of visual concepts as in [23] for Microsoft COCO dataset. For each input image, Faster R-CNN output a set of rectangular object proposals with an objectness score. We sort these scores and select the top-5 areas as the objects detected in this image. Each object then is represented as a vector of 4096 dimensions by the 'fc7' layer of faster R-CNN. Also, the spatial location of each object is represented as a vector of 4096 dimensions by the 'fc7' layer of VGG net. We concatenate these two vectors to form annotation vectors $\{A_i\}$, which yields the dimension of annotation vectors as 8192. We also use the 'fc7' layer of VGG Net to extract features of the whole image and repeat this vector twice. So for each individual input image, the output of the proposed encoding part is a matrix of 6*8192 dimensions.

The proposed decoding part is the LSTM network [7] with attention mechanism. We map each word to a vector of 1000 dimensions and set the hidden layers of LSTM to be 1000 dimensions. We use *tanh* as nonlinear activation function σ. For training, we use stochastic gradient descent with momentum = 0.9 to do model updating with a mini-batch size of 100, we set initial learning rate to 0.01 and decrease it by half after every 20,000 iterations. During testing, we do beam size of 4 to decode words until the end of sentence symbol is reached.

**Table 1.** Results on the MSCOCO dataset.

| Method | Bleu-1 | Bleu-2 | Bleu-3 | Bleu-4 | METEOR | CIDEr | ROUGH-L |
|---|---|---|---|---|---|---|---|
| NIC [8] | 66.6 | 46.1 | 32.9 | 24.6 | - | - | - |
| LRCN [19] | 62.79 | 44.19 | 30.41 | 21 | - | - | - |
| DeepVS [10] | 62.5 | 45 | 32.1 | 23 | 19.5 | 66 | - |
| m-RNN [20] | 67 | 49 | 35 | 25 | - | - | - |
| soft-att [11] | 70.7 | 49.2 | 34.4 | 24.3 | 23.9 | - | - |
| hard-att [11] | 71.8 | 50.4 | 35.7 | 25 | 23.04 | - | - |
| g-LSTM [21] | 67 | 49.1 | 35.8 | 26.4 | 22.74 | 81.25 | - |
| ATT [12] | 70.9 | 53.7 | 40.2 | 30.4 | 24.3 | - | - |
| RA+SF [14] | 69.1 | 50.4 | 35.7 | 24.6 | 22.1 | 78.3 | 50.1 |
| (RA+SF)-BEAM10 [14] | 69.7 | 51.9 | 38.1 | 28.2 | 23.5 | 83.8 | 50.9 |
| Ours | 70.4 | 53.1 | 39.2 | 29 | 23.8 | 85 | 52.1 |

### 3.3 Evaluation Results

Following previous works, we evaluate the proposed model using standard metrics: BLEU-1,2,3,4, METEOR, CIDEr and ROUGH-L (the higher the better), all the metrics are computed by using the codes released by COCO Evaluation Server [18]. The results are shown in Table 1, where "Ours" indicates the performance of the proposed model. Comparing the results of [8, 19, 10, 20], which are CNN-RNN models, with results of [11, 21, 14], which are attention based models, we can find that to encode each image into a set of annotation vectors which represent different subregions of the input image is better and more optimized then to encode the whole image into a single feature vector. Since the proposed model is directly developed from the soft-attention approach [11], therefore, its comparison is highly demanding. The major difference between the proposed model and the soft-attention approach [11] is that we use the information of objects and their spatial information in images for sentence generation. From Table 1, it's clear that the proposed model has achieve better performance than the soft-attention approach [11], which proves that the additional information really help machines to achieve better image understanding. Comparing with the best model which is ATT model [12] up to date, our results are still comparable even though they used a stronger CNN construction which is GoogLeNet instead of VGG16 in our case. Also, we claim that the proposed model takes less computation and more optimized than ATT model since GoogLeNet is more deeper model than VGG16.

## 4   Conclusion

In this paper, we present a multi-model Neural Network that automatically learns to describe the content of images. Our model first extracts the information of objects and their spatial locations in an image, and then a deep recurrent neural network (RNN) based on LSTM units with attention mechanism generates a description sentence. Each word of the description is automatically aligned to different objects in the input image when it is generated. The proposed model is more optimized compared to other benchmark algorithms on the ground that its implementation is totally made on human visual system. We hope that this paper will serve as a reference guide for researchers to facilitate the design and implementation image captioning.

## References

1. Krizhevsky A, Sutskever I, Hinton G E. Imagenet classification with deep convolutional neural networks[C]//Advances in neural information processing systems. 2012: 1097-1105.
2. Simonyan K, Zisserman A. Very deep convolutional networks for large-scale image recognition[J]. arXiv preprint arXiv:1409.1556, 2014.
3. Szegedy C, Liu W, Jia Y, et al. Going deeper with convolutions[C]//Proceedings of the IEEE Conference on Computer Vision and Pattern Recognition. 2015: 1-9.
4. Girshick R, Donahue J, Darrell T, et al. Region-based convolutional networks for accurate object detection and segmentation[J]. IEEE transactions on pattern analysis and machine intelligence, 2016, 38(1): 142-158.
5. Girshick R. Fast r-cnn[C]//Proceedings of the IEEE International Conference on Computer Vision. 2015: 1440-1448.
6. Ren S, He K, Girshick R, et al. Faster r-cnn: Towards real-time object detection with region proposal networks[C]//Advances in neural information processing systems. 2015: 91-99.
7. Hochreiter S, Schmidhuber J. Long short-term memory[J]. Neural computation, 1997, 9(8): 1735-1780.
8. Vinyals O, Toshev A, Bengio S, et al. Show and tell: A neural image caption generator[C]//Proceedings of the IEEE Conference on Computer Vision and Pattern Recognition. 2015: 3156-3164.
9. Mao J, Xu W, Yang Y, et al. Deep captioning with multimodal recurrent neural networks (m-rnn)[J]. arXiv preprint arXiv:1412.6632, 2014.

10. Karpathy A, Fei-Fei L. Deep visual-semantic alignments for generating image descriptions[C]//Proceedings of the IEEE Conference on Computer Vision and Pattern Recognition. 2015: 3128-3137.
11. Xu K, Ba J, Kiros R, et al. Show, attend and tell: Neural image caption generation with visual attention[C]//International Conference on Machine Learning. 2015: 2048-2057.
12. You Q, Jin H, Wang Z, et al. Image captioning with semantic attention[C]//Proceedings of the IEEE Conference on Computer Vision and Pattern Recognition. 2016: 4651-4659.
13. Uijlings J R R, Van De Sande K E A, Gevers T, et al. Selective search for object recognition[J]. International journal of computer vision, 2013, 104(2): 154-171.
14. Jin J, Fu K, Cui R, et al. Aligning where to see and what to tell: image caption with region-based attention and scene factorization[J]. arXiv preprint arXiv:1506.06272, 2015.
15. Bahdanau D, Cho K, Bengio Y. Neural machine translation by jointly learning to align and translate[J]. arXiv preprint arXiv:1409.0473, 2014.
16. Mnih V, Heess N, Graves A. Recurrent models of visual attention[C]//Advances in neural information processing systems. 2014: 2204-2212.
17. Ba J, Mnih V, Kavukcuoglu K. Multiple object recognition with visual attention[J]. arXiv preprint arXiv:1412.7755, 2014.
18. Andrej Karpathy, "neuraltalk2", https://github.com/karpathy/neuraltalk2.
19. Donahue J, Anne Hendricks L, Guadarrama S, et al. Long-term recurrent convolutional networks for visual recognition and description[C]//Proceedings of the IEEE conference on computer vision and pattern recognition. 2015: 2625-2634.
20. Mao J, Xu W, Yang Y, et al. Deep captioning with multimodal recurrent neural networks (m-rnn)[J]. arXiv preprint arXiv:1412.6632, 2014.
21. Jia X, Gavves E, Fernando B, et al. Guiding the long-short term memory model for image caption generation[C]//Proceedings of the IEEE International Conference on Computer Vision. 2015: 2407-2415.
22. Lin T Y, Maire M, Belongie S, et al. Microsoft coco: Common objects in context[C]//European Conference on Computer Vision. Springer International Publishing, 2014: 740-755.
23. Fang H, Gupta S, Iandola F, et al. From captions to visual concepts and back[C]//Proceedings of the IEEE Conference on Computer Vision and Pattern Recognition. 2015: 1473-1482.